\documentclass[sigconf]{acmart}
\AtBeginDocument{%
  \providecommand\BibTeX{{%
    \normalfont B\kern-0.5em{\scshape i\kern-0.25em b}\kern-0.8em\TeX}}}

\copyrightyear{2024}
\acmYear{2024}
\acmDOI{XXXXXXX.XXXXXXX}

\acmConference[HRI'24]{March 11--15, 2024}{Human-LLM Interaction Workshop}{Boulder, CO}

\usepackage{csquotes}

\begin{document}

\title{Large Language Models Enable Automated Formative Feedback in Human-Robot Interaction Tasks}

\author{Emily Jensen}
\affiliation{%
  \institution{University of Colorado Boulder}
  \city{Boulder}
  \state{Colorado}
  \country{USA}}
\email{emily.jensen@colorado.edu}

\author{Sriram Sankaranarayanan}
\affiliation{%
  \institution{University of Colorado Boulder}
  \city{Boulder}
  \state{Colorado}
  \country{USA}}
\email{srirams@colorado.edu}

\author{Bradley Hayes}
\affiliation{%
  \institution{University of Colorado Boulder}
  \city{Boulder}
  \state{Colorado}
  \country{USA}}
\email{bradley.hayes@colorado.edu}
\renewcommand{\shortauthors}{Jensen, et al.}




 \begin{teaserfigure}
   \includegraphics[width=\textwidth]{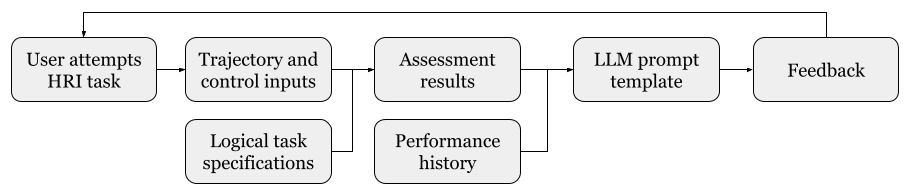}
   \caption{Example pipeline for generating automated feedback for HRI tasks}
   \label{fig:pipeline}
 \end{teaserfigure}

\settopmatter{printacmref=false}
\maketitle

\section{Introduction}
Rapid advances in robotic systems will significantly impact the future of work. Workers will need to be able to quickly adapt to using new systems as new capabilities and application domains emerge. A recent report estimates that one third of job requirements will require technological skills that are not yet considered crucial~\cite{schwab_future_2020}. It is also estimated that 50\% of existing employees will need to be retrained or upskilled by 2025 to keep up with technological advancement, placing significant pressure on both employers and employees to meet these demands~\cite{schwab_future_2020, li_reskilling_2022}. As such, we need scalable methods to train users for human-robot interaction (HRI) tasks.

Automated feedback is a promising approach to scale up training for HRI tasks. By pairing domain knowledge representations with effective assessment, automated feedback systems can identify a learner's current strengths and weaknesses and suggest future actions that will help the learner master the target task. In this paper, we discuss how large language models (LLMs) can be used as a tool for providing automated feedback for learning HRI tasks alongside illustrative examples.

\section{Position Statement}
Representing knowledge and assessing someone's ability in an HRI task is difficult, due to complex objectives and high variability in human performance. In \cite{jensen_more_2023}, we begin to address this question by breaking down HRI tasks into objective primitives that can be combined sequentially and concurrently (e.g., maintain slow speed and reach waypoints). We then show that signal temporal logic specifications, paired with a robustness metric, are a useful tool for assessing performance along each primitive. These formal methods allow designers to precisely represent ideal trajectories. This formulation admits explainability, as one can identify and elaborate upon specific objectives that learners did not accomplish.

We claim that \textit{LLMs can be paired with formal analysis methods to provide accessible, relevant feedback for HRI tasks}. While logic specifications are useful for defining and assessing a task, these representations are not easily interpreted by non-experts. Luckily, LLMs are adept at generating easy-to-understand text that explains difficult concepts. By integrating task assessment outcomes and other contextual information into an LLM prompt, we can effectively synthesize a useful set of recommendations for the learner to improve their performance (see Figure~\ref{fig:pipeline} as one example).

\section{Challenges and Opportunities}
To the best of our knowledge, there has been no systematic work addressing how to train humans to effectively work with robotic systems. If we consider training motor skills more broadly, many approaches focus on augmenting sensory input to provide control-level feedback during a task \cite{sigrist_augmented_2013}. Some examples include visualizing the predicted future trajectory of a drone \cite{wilde_predictive_2020} or generating a haptic response to bias the operator to an ideal course \cite{agrawal_toddlers_2012, rognon_haptic_2019}. These approaches to feedback fall short in several respects: (1) just-in-time feedback does not promote intentional reflection from the learner on how to improve their performance, (2) feedback is not integrated with an established training curriculum, and (3) feedback does not adapt to a learner's learning trajectory over time.

\begin{figure}
    \centering
    \includegraphics[width=\linewidth]{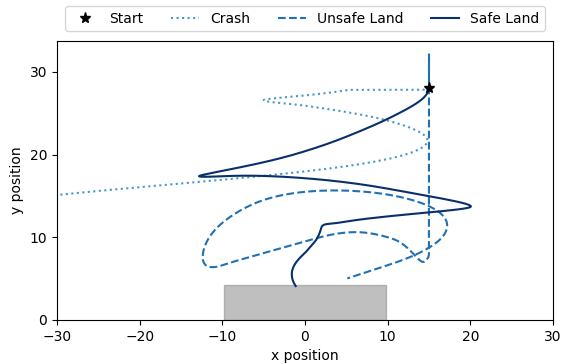}
    \caption{Example of three trajectories in a 2D drone landing task. The pilot must keep the drone within the black boundary and safely reach the gray landing pad with a velocity $\leq 5~m/s$ and attitude $\leq 10^{\circ}$.}
    \label{fig:trajectory}
    \vspace{-20 pt}
\end{figure}

LLMs are a promising technology poised to tackle these challenges in generating feedback. To address challenge (1), we can develop feedback templates that include elements of effective formative feedback. Formative feedback is an established approach in education that focuses on motivating learners, having them reflect on their performance, and providing a manageable amount of feedback they can use on their next attempt. For example, we generated the following feedback for the safe trajectory shown in Figure~\ref{fig:trajectory}:

\begin{quote}
Great job maintaining the safety boundaries throughout your flight! While your landing was accurate with respect to y-axis, it seems there's room for improvement in maintaining the x-axis positioning within the landing pad. To refine this, consider adjusting the roll input slightly when approaching the landing pad. Reflect on the trajectory taken during the descent and think about how you could make it more centered over the pad. I'm confident that with a few tweaks, you'll nail the perfect landing next time! 
\end{quote}

LLMs can also be integrated with broader training systems that document domain knowledge and skill structures, addressing challenge (2). Finally, LLMs can access historical records of the learner's performance to understand repeated mistakes and opportunities for growth. The greatest strength of LLMs is generating friendly, approachable text, making it ideal for providing feedback that learners perceive positively.

\section{Potential Impact}

LLM feedback is easy to iterate on and integrate into existing technical workflows. For example, if a task involves multiple robots, we can quickly modify a template prompt for feedback by including a short description of each robot's dynamics or the specific part of the task it is used for. If we need a new feedback format to test a new learning theory, we can swap out the part of the prompt that tells the LLM how to frame its response. Changing the feedback presentation is also straightforward; the LLM can act as an intermediate interface between the task and feedback, generating appropriate low-level code to display on a virtual reality headset or to be spoken by a social robot. 

The social implications of the proposed approach are also worth considering. As more people are required to interact with robots, training needs to be scalable and personalized to each learner. LLMs can help us reach this goal by making training a friendlier and more appealing process. When paired with commercially available products like virtual reality headsets, training for robotic systems can be accessible to a more diverse group of learners and enable them to train for technical jobs. Additionally, automated feedback generation lessens the burden on human instructors, who will not need to provide as much direct oversight during training.
\vspace{-7 pt}

\section{Challenges}
As with any new technology that is not fully understood, there are many questions to consider before integrating LLM feedback into high-stakes or high-impact HRI domains. 

The first consideration is anticipating and handling unexpected outputs from the feedback system. It is widely known that LLMs can produce incorrect and harmful responses due to the stochastic nature of the model, biases in the training data, and nuances in system prompts. An automated feedback system should have internal processes to moderate potential harmful outputs, which can be built into the prompt. For example, Tree-of-Thought prompting \cite{tree-of-thought-prompting} can be used to emulate experts giving multiple feedback variations and having them reach a consensus based on internal knowledge (recent performance, historical errors, possible emotional states). This approach allows the LLM to recognize and discard inaccurate or poorly phrased feedback. Additional research can collaborate with natural language processing efforts to develop safety alignment when training new models \cite{dong2024attacks}.

The second consideration is implementing the feedback system in a robust and sustainable manner. A feedback system requires a thorough yet flexible knowledge representation of the target domain. Using principles from participatory design \cite{spinuzzi2005methodology}, system creators and domain experts can work together to identify key learning outcomes and assessment criteria. These core learning concepts can then be codified in a formal framework such the Knowledge-Learning-Instruction Framework \cite{koedinger_knowledgelearninginstruction_2012}, which associates each practice item with one or more knowledge components. Finally, feedback systems should operationalize theory-driven intervention strategies for providing feedback, such as the Zone of Proximal Development \cite{vygotsky_mind_1978} or Deliberate Practice \cite{Ericsson1993}. This would bring more rigor to HRI studies while also contributing to interdisciplinary discourse on training for real-world domains.


\section{Call to Action}
In this paper we have presented concepts suggesting that LLMs are a promising tool for automatically generating feedback for learning HRI tasks. By using theory-driven approaches to assessing learner performance and providing feedback, the HRI community can develop robust systems that both contribute to a larger discussion of learning and training and fill critical gaps in domains with immediately relevant applications and societal benefits.

As this area develops, we propose a few starting research directions. First, we must understand how learning for HRI tasks (which commonly involve physical manipulation or inter-agent communication) differs from the traditional classroom context and how this impacts theories of learning. Second, we can investigate how learner agency fits into an automated feedback system; how much control do learners want over the tone, content, and delivery of their feedback? Do these choices make the feedback less effective? Finally, we need to consider the temporal aspect of learning. How can feedback address both the most recent learning attempt and glean long-term trends from historical data? These directions can prompt more nuanced questions as the research develops.

\bibliographystyle{ACM-Reference-Format}
\bibliography{sample-base}

\end{document}